\documentclass[sigconf]{acmart}
\usepackage[acronym]{glossaries}
\newacronym{hci}{HCI}{Human-Computer Interaction}
\newacronym{gui}{GUI}{Graphical User Interfaces}
\newacronym{nui}{NUI}{Natural User Interface}
\newacronym{cnn}{CNN}{Convolutional Neural Network}
\newacronym{rgbd}{RGBd}{Red, Green, Blue and depth}
\newacronym{rgb}{RGB}{Red, Green and Blue}
\newacronym{roi}{RoI}{Region of Interest}
\newacronym{ml}{ML}{Machine Learning}
\newacronym{opencv}{OpenCV}{Open Computer Vision}

\usepackage{orcidlink}
\usepackage{graphicx,wrapfig,lipsum, caption, subcaption}
\AtBeginDocument{%
  \providecommand\BibTeX{{%
    \normalfont B\kern-0.5em{\scshape i\kern-0.25em b}\kern-0.8em\TeX}}}

\copyrightyear{2022} 
\acmYear{2022} 
\setcopyright{acmlicensed}\acmConference[PETRA '22]{The15th International
Conference on PErvasive Technologies Related to Assistive Environments}{June
29-July 1, 2022}{Corfu, Greece}
\acmBooktitle{The15th International Conference on PErvasive Technologies
Related to Assistive Environments (PETRA '22), June 29-July 1, 2022, Corfu,
Greece}
\acmPrice{15.00}
\acmDOI{10.1145/3529190.3534749}
\acmISBN{978-1-4503-9631-8/22/06}

\begin{document}

\title{Real-Time Gesture Recognition with Virtual Glove Markers}

\author{Finlay McKinnon}
\affiliation{%
  \institution{School of Science and Technology\\Nottingham Trent University}
  \streetaddress{Clifton Lane}
  \city{Nottingham}
  \country{United Kingdom}
  \postcode{NG11 8NS}}

\author{David Ada Adama\orcidlink{0000-0002-2650-857X}}
\authornotemark[1] 
\affiliation{%
  \institution{School of Science and Technology\\ Nottingham Trent University}
  \streetaddress{Clifton Lane}
  \city{Nottingham}
  \country{United Kingdom}
  \postcode{NG11 8NS}
}
\email{david.adama@ntu.ac.uk}

\author{Pedro Machado\orcidlink{0000-0003-1760-3871}}
\affiliation{%
  \institution{School of Science and Technology\\Nottingham Trent University}
  \streetaddress{Clifton Lane}
  \city{Nottingham}
  \country{United Kingdom}
  \postcode{NG11 8NS}}
\email{pedro.machado@ntu.ac.uk}

\author{Isibor Kennedy Ihianle\orcidlink{0000-0001-7445-8573 }}
\affiliation{%
  \institution{School of Science and Technology\\Nottingham Trent University}
  \streetaddress{Clifton Lane}
  \city{Nottingham}
  \country{United Kingdom}
  \postcode{NG11 8NS}}
\email{isibor.ihianle@ntu.ac.uk}

\renewcommand{\shortauthors}{McKinnon, et al.}

\begin{abstract}
  Due to the universal non-verbal natural communication approach that allows for effective communication between humans, gesture recognition technology has been steadily developing over the previous few decades. Many different strategies have been presented in research articles based on gesture recognition to try to create an effective system to send non-verbal natural communication information to computers, using both physical sensors and computer vision. Hyper accurate real-time systems, on the other hand, have only recently began to occupy the study field, with each adopting a range of methodologies due to past limits such as usability, cost, speed, and accuracy. A real-time computer vision-based human-computer interaction tool for gesture recognition applications that acts as a natural user interface is proposed. Virtual glove markers on users hands will be created and used as input to a deep learning model for the real-time recognition of gestures. The results obtained show that the proposed system would be effective in real-time applications including social interaction through telepresence and rehabilitation.
\end{abstract}


\maketitle
\section{Introduction}\label{sec:intro}
Hand gesture recognition is an important area of development and a focus point for \gls{hci} that can be applied to a variety of areas, including sign language recognition, virtual and augmented reality, assisted living technology, and industrial use \cite{Wang2009ColourGlove, HanMEgATrack, VarunDisables2019, Bangaru2020}.

Gesture recognition may be viewed as a promising research topic when discussing \gls{hci} and computer vision, due to the innate properties of gestures, being natural and decisive forms of communication. In \gls{hci} research, it is important to consider the amount of usability friction that the human has when interacting with the computer. A device with many physical components would have a considerably high usability friction, as layers of complexity are being added to the interaction from the user's perspective. The use of alternative approaches are desired for less complexity. For example, in applications involving gesture recognition, the use of a vision-based device would have an incredibly low usability friction, as the movements are natural to the user, and there are no additional layers.

Over the years, gesture \gls{hci} has been improved upon to a great amount, starting with unsatisfactory results using bulky physical sensors \cite{Zimmerman1986}, to accurate recognition using everyday objects \cite{Porzi2013}. However, the development of optical recognition technology solutions has been slow, due to inconsistencies in the nature of optical sensing and how it is affected by background and environmental lighting, occlusions, processing times against image resolution and frame rates all of which makes gesture recognition performance sub-optimal \cite{Murthy_areview}.

With regard to the wide range of hand gesture recognition applications, the technology has been applied to many domains. Hand gesture recognition can be used for a safer driving experience to minimise driver distraction \cite{MolchanovDrivingGesture}, aiding the visually impaired via smart-watch gesture recognition \cite{Porzi2013}, and construction worker safety training \cite{Bangaru2020}. Such technology has also been suggested for highly precise situations, as being used for augmented reality \gls{gui} dance in robotic surgery \cite{WenSurgery2010}, as well as social interactions with its suggested use being included in an autonomous telepresence robot for remote conferencing \cite{DoInteraction2013}. With these approaches, it can be understood that the technology holds promise in a variety of fields, aiding in vastly different areas with multiple differing methods to yield beneficial results. 

According to Donchysts et al. \cite{Donchyts2014}, in order to truly capitalise on the extent of gesture recognition, the user's experience should be as fluid as possible, which can be achieved with low usability friction that computer-vision based gesture recognition can offer as capable through natural user interfaces, which traditional means of interaction cannot. Traditional methods of \gls{hci}, command lines, \gls{gui}, keyboards and mice are all inconvenient and unnatural \cite{Murthy_areview}, and may even be unusable if the user has some form of physical impairment. 

A form of human computer interaction which is a \gls{nui} through a real-time computer vision-based hand gesture recognition system is proposed. It takes advantage of the benefits of hand gesture recognition as a form of \gls{hci} over the traditional methods highlighted above. The research uses a means of image processing to extract regions of a user's hand which are used as virtual glove markers for detecting specific points of the hand.

The remainder of the paper is structured as follows; Section \ref{sec:review} presents a review of related work. In Section \ref{sec:methodology}, the methodology of the proposed real-time gesture recognition with virtual glove markers system is discussed. Section \ref{sec:exp} describes the experimental setup and results obtained. Section \ref{sec:conclusion} concludes the paper and gives directions for future work.

\section{Related Work}\label{sec:review}
Research into gesture recognition has been an important research area due to its wide-ranging areas of application, some of which includes solutions to minimise distractions and dangers while driving \cite{MolchanovDrivingGesture} to comfort and efficiency \cite{ReifingerAR2007}. A brief background research about gesture recognition is presented in this section.

Recent advances have shown the different approaches to hand gesture recognition. Stergiopolou and Papamarkos \cite{Stergiopoulou2006} proposed colour segmentation and a neural gas neural network to identify finger positioning and gestures utilising histograms. Shangchen et al. \cite{HanMEgATrack} proposed a robust system utilising a neural network for hand detection accompanied by hand key-point location estimations. Yin and Xie \cite{YinPostureSeg2007} suggests applying an RCE neural network for hand segmentation, 2D and 3D feature extraction to reconstruct hand movements and gestures for \gls{hci}. Feature extraction based on colour segmentation, depth information, or deep learning, as well as feature segmentation, key point estimation, and histograms to estimate hand position and orientation, are frequently proposed approaches for developing a vision-based hand gesture recognition system; however, these are only a few of the many proposed methods. Recognition based on skin colour, appearance based on backdrop subtraction, motion-based, depth-based, and deep-learning based, as well as a variety of additional techniques are covered in \cite{Oudah2020}.

More recently, in 2020, driver distraction was responsible for roughly $10\%$ of motor vehicle deaths \cite{TrafficSafety2020}, though with the research proposed by Molchanov \cite{MolchanovDrivingGesture}, driver distraction could be lowered through changing the \gls{hci} method within cars to a gesture based system. Furthermore, research done by Reifinger \cite{ReifingerAR2007} shows that gesture based \gls{hci} can be more intuitive, more comfortable, and up to 60$\%$ quicker than the aforementioned traditional forms of \gls{hci}. 

The cost of physical sensors, complexity of gesture recognition and hand tracking algorithms, presents challenges and limitations to the accessibility of the aforementioned forms \gls{nui} to users; however, with the approach proposed in this paper, it is possible to create relatively accurate recognition with hardware cheap and accessible to potential users. 

The approach utilising a virtual glove was inspired by the classical physical glove marker approach demonstrated by \citep{Wang2009ColourGlove}. The simplicity of the core idea, utilising colour markers to indicate sections of a hand and machine learning methods to approximate the hand's positioning. This approach adapts the improved machine learning algorithm proposed by Wang et al. \citep{Wang2009ColourGlove} which allows for efficient and accurate algorithms to assist in applying a virtual glove to the user's hand in real time. 

\begin{figure}[t]
    \centering
    \includegraphics[width=8cm]{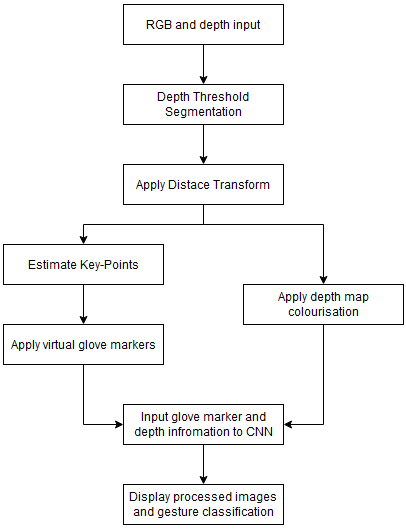}
    \caption{Overview of real-time gesture recognition system using virtual glove markers.}
    \label{fig:simplifiedProcessDiagram}
\end{figure}

\section{Methodology}\label{sec:methodology}
The aim of this paper is to provide insight into an approach to creating an \gls{nui} through the use of computer vision, specifically utilising depth information, segmentation and a \gls{cnn}. This will be accomplished by the creation of virtual glove markers, as opposed to the standard physical markers as proposed by Wang et al. \citep{Wang2009ColourGlove, ReifingerAR2007}. The research provides a mixture of techniques not commonly used together to attempt to create a system in which static hand gestures may be accurately understood by the system as a new promising form of \gls{hci} in the form of a \gls{nui} using glove markers. Glove markers allow a vision-based tracking system to extract the orientation and position of the hand. However, the method requires physical gloves to be worn as a medium for the system to work. An alternative method that includes replacing the physical glove marker technique into virtual glove markers via segmenting the hand after key-point estimation is proposed in this article.

\begin{figure}[t]
  \begin{subfigure}[b]{4cm}
    \includegraphics[width=4cm]{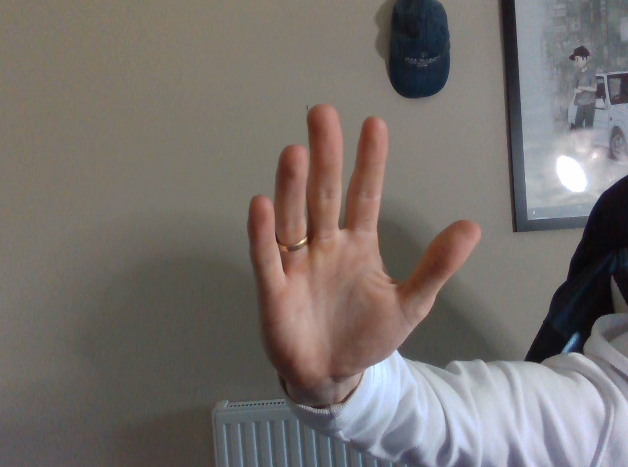}
    \caption{}
  \end{subfigure}
  \hfill
  \begin{subfigure}[b]{4cm}
    \includegraphics[width=3.7cm]{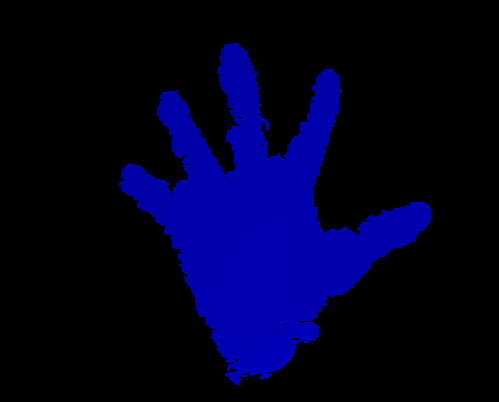}
    \caption{}
  \end{subfigure}
  \hfill
  \begin{subfigure}[b]{4cm}
    \includegraphics[width=4cm]{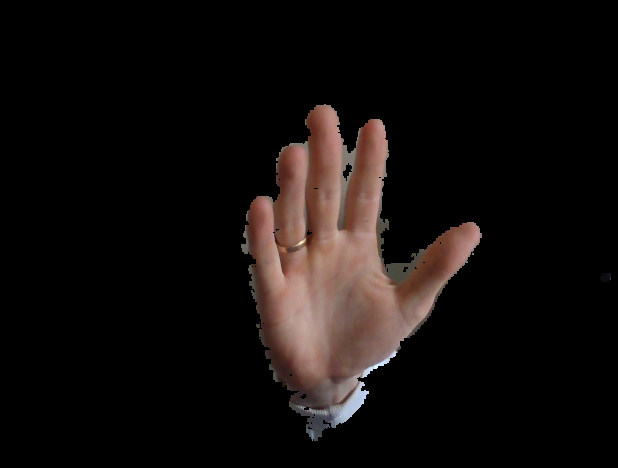}
    \caption{}
  \end{subfigure}
  \caption{Depth threshold filter applied to an \gls{rgb} image. (a) \gls{rgb} image input. (b) Colourised depth input from sensor. (c) \gls{roi} segmentation through depth threshold.}
  \label{fig:ExampleDepthTreshold}
\end{figure}

An overview of the methodology for the real-time gesture recognition system using virtual glove markers presented in this paper is given in Figure \ref{fig:simplifiedProcessDiagram}. The key steps in the methodology are summarised as follows:

\textbf{\gls{rgbd} input:} Firstly, the research makes use of \gls{rgbd} information of the hand, obtained using a depth sensor. An example of both \gls{rgb} and depth input frames can be seen in Figures \ref{fig:ExampleDepthTreshold}(a) and \ref{fig:ExampleDepthTreshold}(b).
    
\textbf{Depth Thresholding:} in the system is the act of segmenting the \gls{roi}, being the user's hand in this case, from the rest of an image. This will be utilised as form of reducing the computational workload when the image is passed to the recognition model, as there is less nonessential data in each image to be processed. From the collected data, the \gls{rgb} frame will undergo a depth-threshold transformation via the depth data obtained in the respective depth frame. This depth-threshold transformation will effectively remove all \gls{rgb} data from a frame if the information is past a given distance, being the threshold. Figure \ref{fig:ExampleDepthTreshold}(b) shows an example of image after depth thresholding has been applied. The threshold chosen for the research is 500mm from the camera, used in an environment where there are no objects closer than 500mm other than the user's hand.
    
\begin{figure}[t]
  \begin{subfigure}[b]{4cm}
    \includegraphics[width=4cm]{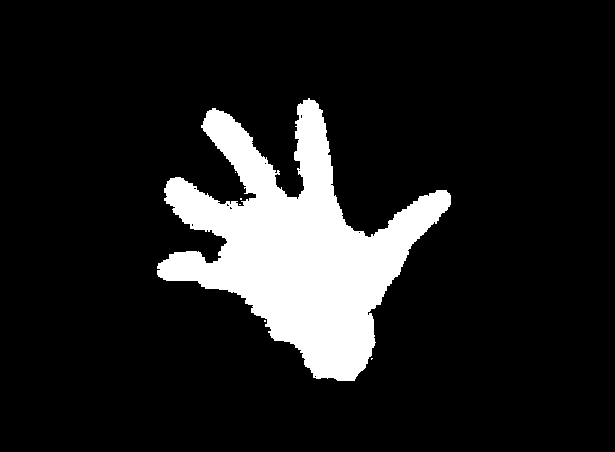}
    \caption{}
  \end{subfigure}
  \hfill
  \begin{subfigure}[b]{4cm}
    \includegraphics[width=4cm]{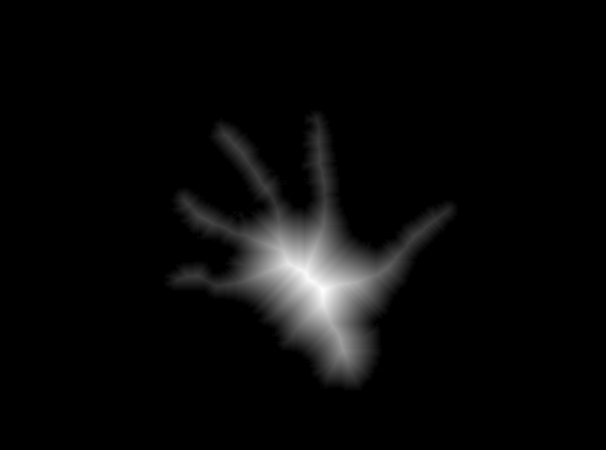}
    \caption{}
  \end{subfigure}
  \caption{Distance transform applied to segmented binary image. (a) Binary conversion of the \gls{roi} segmented image. (b) Binary image after distance transform.}
  \label{fig:VisualExampleDistanceTransform}
\end{figure}

\textbf{Distance Transform:} the \gls{rgb} image processed in the \gls{roi} segmentation stage, is converted to binary image by a transformation of the \gls{rgb} image matrix, converting the image to a $2D$ binary image, with a value of $0$ representing background and value $1$ representing the hand as shown in Figure \ref{fig:VisualExampleDistanceTransform}(a). A distance transform is then applied to the binary image of the hand, being used to calculate a representation of the most central point in the hand. This is possible by calculating the relative distance of each hand pixel element from a background pixel element as shown in Figure \ref{fig:VisualExampleDistanceTransform}(b).
    
\textbf{Key-Point (Finger Point) Estimation:} involves the use of a \gls{ml} algorithm known as MediaPipe\footnote{Available online, \protect\url{https://ai.googleblog.com/2019/08/on-device-real-time-hand-tracking-with.html}, last accessed: 21/03/2022} to estimate the fingertips and knuckles in the given frame. MediaPipe creates key points on the hand known as landmarks, these landmarks including the fingertips and knuckle points, allow the system to efficiently estimate their location. MediaPipe has been used in this system as a tool for proof of concept, due to computational constraints.
The fingertip key point coordinates of the frame and then saved for later use.

\textbf{Centre Palm Point Estimation:} as the machine learning algorithm utilised for finger key-points does not calculate a central palm point, the system utilises the distance transform to calculate the most central point of the hand. This has been done through searching for the highest value in the distance transform, representing the pixel furthest from a background pixel, which on a hand is generally the central point of the palm. The radius of the palm can be calculated by the value of the pixel element, as the value increments by one per pixel, this will give an accurate relative distance to the edges of the palm. This estimated palm area is then visually displayed on the virtual glove, using the central point and calculated radius of the palm. A moving average, of the previous 5 frames, is used to smooth the jitter.

\begin{figure}{t}
    \centering
    \includegraphics[width=4cm]{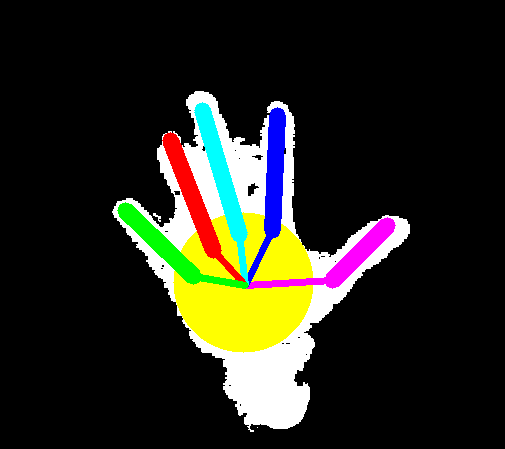}
    \caption{Processed hand depth image with virtual glove marker.}
    \label{fig:virtualGloveMarker}
\end{figure}

The virtual glove is then drawn onto the hand, atop the processed image with the estimated palm drawn to it. The proposed glove has been applied through plotting lines using the \gls{opencv} library\footnote{Available online, \protect\url{https://opencv.org/}, last accessed 21/03/2022} to points gathered from the key-point estimation process, creating palm to knuckle then knuckle to fingertip links as shown in Figure \ref{fig:virtualGloveMarker}. The processed image displaying each portion of the hand in a distinct colour will act as the virtual glove to be used in the model for gesture categorisation.

\textbf{Gesture Categorisation using \gls{cnn}:} the categorisation of each gesture shall be done by passing each frame to a \gls{cnn}, categorising each frame based on previous training of specific images of predefined gestures. The \gls{cnn} will be trained on both the depth information and virtual glove, the data being given categories to recognise different gestures as.

\section{Experimental Results}\label{sec:exp}
\begin{figure}[t]

          \begin{subfigure}[b]{3.5cm}
            \includegraphics[width=3.5cm]{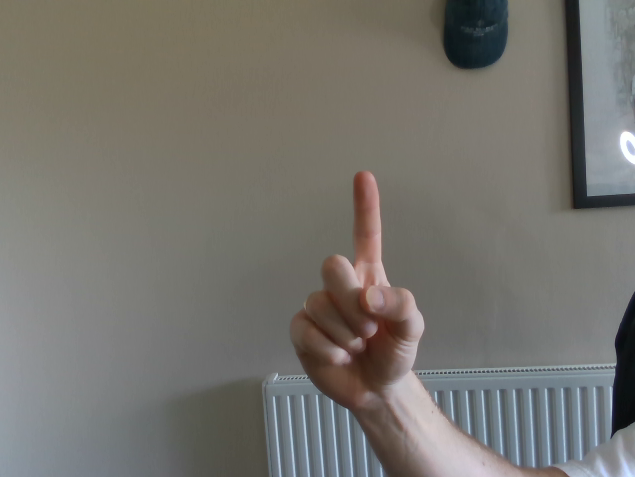}
            \caption{}
            \label{fig:RGBOneFinger}
          \end{subfigure}
          \hfill
          \begin{subfigure}[b]{3.5cm}
            \includegraphics[width=3.5cm]{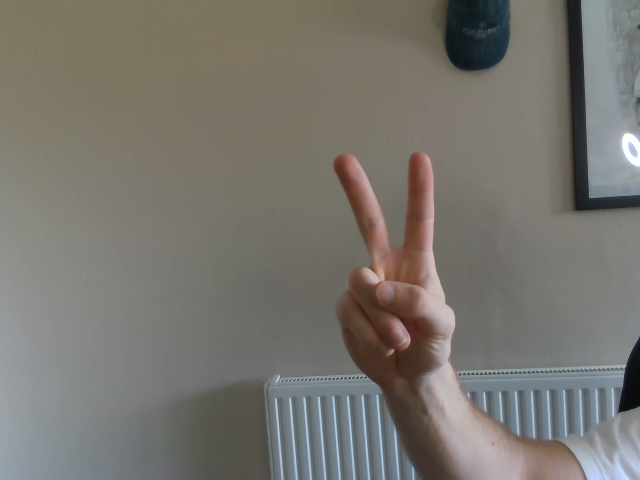}
            \caption{}
            \label{fig:RGBTwoFinger}
          \end{subfigure}
          
          \begin{subfigure}[b]{3.5cm}
            \includegraphics[width=3.5cm]{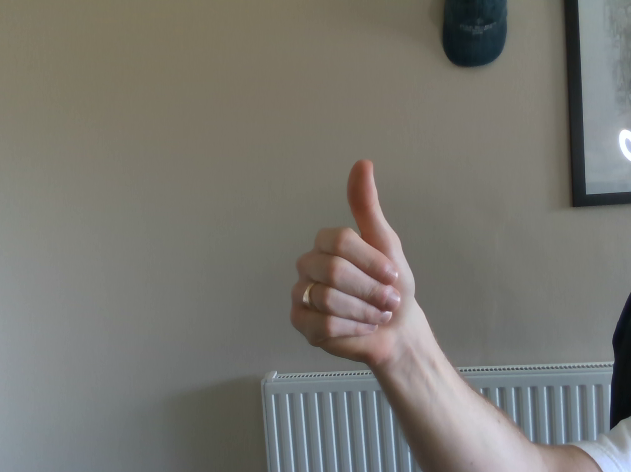}
            \caption{}
            \label{fig:RGBThumb}
          \end{subfigure}
          \hfill
          \begin{subfigure}[b]{3.5cm}
            \includegraphics[width=3.5cm]{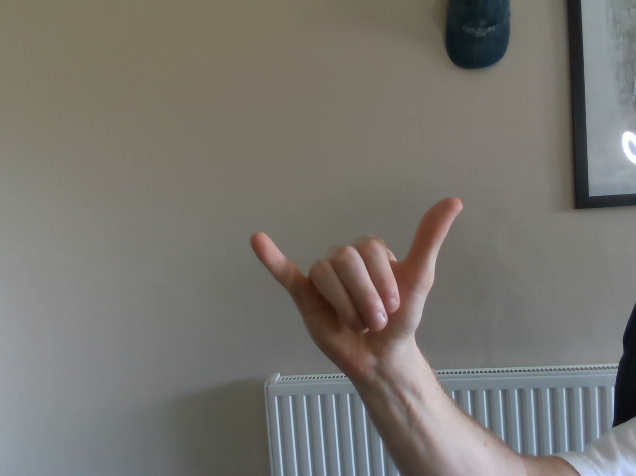}
            \caption{}
            \label{fig:RGBShaka}
          \end{subfigure}
          
          \begin{subfigure}[b]{3.5cm}
            \includegraphics[width=3.5cm]{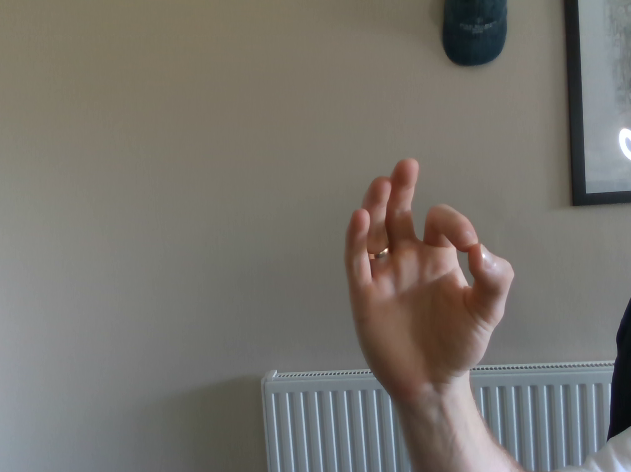}
            \caption{}
            \label{fig:RGBOK}
          \end{subfigure}
          
          \hfill
     \caption{\gls{rgb} image of recognised gestures used in this paper. (a) `One Finger' Gesture. (b) `Two Fingers' Gesture. (c) `Thumb' Gesture. (d) `Shaka' Gesture. (e) `OK' Gesture.}
     \label{fig:RGBExamples}
\end{figure}

The experiments conducted to validate the proposed approach to real-time gesture recognition using virtual glove markers utilises an Intel RealSense $D435i$ depth camera \footnote{Available online, \protect\url{https://store.intelrealsense.com/buy-intel-realsense-depth-camera-d435i.html},last accessed: 21/03/2022}. Using this device, both \gls{rgb} and depth information are obtained. Five different gestures are used for the experiments conducted in this paper. These gestures can be seen in Figure \ref{fig:RGBExamples} and comprise of a `one finger', `thumbs-up', `Ok', `two fingers' and `shaka' gestures.

Data was collected for all five gestures. A total of $1500$ frames were collected which corresponds to $300$ frames for each gesture. This data was then pre-processed and fed into the \gls{cnn} model following the methodology discussed in Section \ref{sec:methodology}. Table \ref{table:AccuracyResults} shows the performance of the proposed system on the dataset. This shows the correctly classified gestures, the total number of each gesture and percentage accuracy.

\begin{table*}[] \caption{Real-time accuracy results.}
        \label{table:AccuracyResults}
    \begin{center}
        \begin{tabular}{|c|c c c c c ||c|} 
         \hline
         - & One Finger & Two Finger & Thumb & Shaka & OK & Total \\ [0.5ex]
         \hline
         Correct & 290 & 271 & 270 & 296 & 298 & 1425 \\ \hline
         Attempted & 300 & 300 & 300 & 300 & 300 & 1500   \\ \hline
         Accuracy & 96.67\% & 90.34\% & 90.0\% & 93.3\% & 99.3\% & 95\%  \\ 
         \hline
        \end{tabular}
    \end{center}
\end{table*}

\begin{table*}[]\caption{Comparison of the proposed system with other related works. {\*}BG - Background}  \label{table:UsabilityResults}
    \begin{center}
    \begin{tabular}{|c|c c|c c|} 
     \hline
     - & Training & Validation & Static BG & Dynamic BG  \\ [0.5ex] 
     \hline
     Proposed System & 98.09\% & 98.42\% & 95\% & 95\%  \\ \hline
     Parelli \cite{10.1007/978-3-030-66096-3_18}  & -  & -  & 90\%  & 90\%       \\ \hline
     Chung \cite{ChungDCNNHandGesture} & 99.9\% & 98.1\% & - & - \\ \hline
      Bao \cite{Bao2017TinyHG} & -  & - & 85\% & 85\%\\ \hline
    \end{tabular}
    \end{center}
\end{table*}

Information in the accuracy table displays the overall accuracy of the current model's recognition ability, scoring an overall accuracy of $95\%$. The accuracy of each gesture show some inconsistency, this possibly being due to overfitting of the \gls{cnn} in some aspects; this can be viewed through the difference in accuracy between gestures such as `OK', scoring $99.3\%$ accuracy, and 'Thumb', scoring only $90\%$ accuracy.

The proposed system's average processing speed clocks in at $135ms$, which is significantly less than the average human's ability to detect visual stimuli at between $180ms$ and $200ms$ \cite{JainReaction}. 
The results presented in Table \ref{table:AccuracyResults} yielding total of $95\%$ real time accuracy shows the potential of the proposed system for use in varying applications. Thu supporting the aim to create a system that is easier for users to interact with, being a \gls{nui} and innately being a more natural form of \gls{hci}, based on the results.

To evaluate the system proposed in this paper, a comparison is made with other similar works. This comparison is reported in Table \ref{table:UsabilityResults}. The accuracy of the proposed system demonstrates the functionality of a real-time gesture recognition, which research in this field often lag behind; moreover, statistical comparisons between the proposed system and other systems of similar design proposing non-real time testing results can be made. The system proposed in \cite{ChungDCNNHandGesture} presents a similar system to the work in this paper, achieving a $99.9\%$ accuracy in training and a $95\%$ accuracy in testing, which is very similar to the results obtained in this paper of $98.1\%$ and $95\%$ respectively. However, the proposed system allows for multiple more use cases in dynamic and cluttered environments, as displayed by the accuracy of a more similar system proposed in \cite{Bao2017TinyHG} scoring $85\%$ in dynamic background testing.

Also, the work in \cite{10.1007/978-3-030-66096-3_18} proposed another system similar to what has been proposed in this paper, utilising estimated skeletal points to infer human gestures. The research reported information about real-time complex background testing, scoring a $90\%$ accuracy within these tests which is significantly lower than the results achieved in this paper.

\section{Conclusion and Future Work}\label{sec:conclusion}
This paper proposed a system which is relatively unique in its design, utilising virtual glove markers and \gls{rgbd} data fed to a \gls{cnn} model for real-time gesture recognition. As stated in the introduction, this process has yielded promising results for the computer vision gesture recognition field, though future could be done to improve on the systems' robustness to more gestures.

The proposed system could be improved to make the approach more robust and offset the limitations of the current system. Firstly, there are limitations within the proposed system, being that the depth-threshold algorithm is not entirely accurate, capturing some background around the hand. This may be improved by adding filters to the algorithm to lessen the unwanted sections of the \gls{roi} captured. Furthermore, lighting must be considered with the system as the key-point estimation \gls{ml} algorithm utilises feature detection, the hand must be well enough illuminated to allow the algorithm to accurately find these points.

Due to processing speeds the user may not have immediate recognition of their gesture and must hold the gesture for about $135ms$, this may also mean that gestures done quickly may be missed.

\bibliographystyle{ACM-Reference-Format}
\bibliography{petra22ref}

\end{document}